\documentclass[runningheads]{llncs}
\usepackage{graphicx}
\usepackage{amsmath}
\usepackage{booktabs}
\usepackage{subcaption}
\usepackage{xcolor}
\usepackage{hyperref}
\usepackage{wrapfig}
\usepackage{adjustbox}
\usepackage{amssymb}
\usepackage{pifont}
\newcommand{\argmin}[1]{\underset{#1}{\operatorname{arg}\,\operatorname{min}}\;}

\begin{document}
\title{Domain Adaptive Relational Reasoning for 3D Multi-Organ Segmentation}
%
%
\author{Shuhao Fu\inst{1} \and
Yongyi Lu\inst{1} \and
Yan Wang\inst{1} \and
Yuyin Zhou\inst{1} \and
Wei Shen\inst{1} \and \\
Elliot Fishman\inst{2} \and
Alan Yuille\inst{1}}

\authorrunning{S. Fu et al.}
%
\institute{$^{1}$Johns Hopkins University $^{2}$Johns Hopkins University School of Medicine \\
{\tt\small \{fushuhao6, yylu1989, wyanny.9, zhouyuyiner, shenwei1231\}@gmail.com}\quad
{\tt\small efishman@jhmi.edu}\quad
{\tt\small alan.l.yuille@gmail.com}\\
}
\maketitle              
\begin{abstract}
In this paper, we present a novel unsupervised domain adaptation (UDA) method, named Domain Adaptive Relational Reasoning (DARR), to generalize 3D multi-organ segmentation models to medical data collected from different scanners and/or protocols (domains). Our method is inspired by the fact that the spatial relationship between internal structures in medical images is relatively fixed, \emph{e.g.}, a spleen is always located at the tail of a pancreas, which serves as a latent variable to transfer the knowledge shared across multiple domains. We formulate the spatial relationship by solving a jigsaw puzzle task, \emph{i.e.}, recovering a CT scan from its shuffled patches, and jointly train it with the organ segmentation task. To guarantee the transferability of the learned spatial relationship to multiple domains, we additionally introduce two schemes: 1) Employing a super-resolution network also jointly trained with the segmentation model to standardize medical images from different domain to a certain spatial resolution; 2) Adapting the spatial relationship for a test image by test-time jigsaw puzzle training. Experimental results show that our method improves the performance by $29.60\%$ DSC on target datasets on average without using any data from the target domain during training.

\keywords{Unsupervised domain adaptation  \and Relational reasoning \and Multi-organ segmentation.}
\end{abstract}

\section{Introduction}
Multi-organ segmentation in medical images, {\em{e.g.}}, CT scans, is a crucially important step for many clinical applications such as computer-aided diagnosis of abdominal disease. With the surge of deep convolutional neural networks (CNN), intensive studies of automatic segmentation methods have been proposed. 
But more evidence pointed out the problem of performance degradation when transferring domains \cite{chen2020unsupervised}{\em{e.g.}}, testing and training data come from different CT scanners or suffer from a high deviation of scanning protocols between clinical sites.
For example, training a well-known V-Net \cite{vnet} on our in-house dataset and directly testing it on a public MSD spleen dataset \cite{Simpson2019} yields 43.12\% performance drop in terms of DSC. The reason is that their reconstruction and acquisition parameters are different, {\em{e.g.}}, pitch/table speeds are 0.55-0.65/25.0-32.1 for the in-house dataset, and 0.984–1.375/39.37–27.50 for MSD spleen dataset.
In the context of large-scale applications, generalization capability to deal with scans acquired with different scanners or protocols ({\em{i.e.}}, different domains) as compared to the training data is desirable for machine learning models when deploying to real-world conditions.

\begin{figure}[t]
\centering
\includegraphics[width=0.6\linewidth]{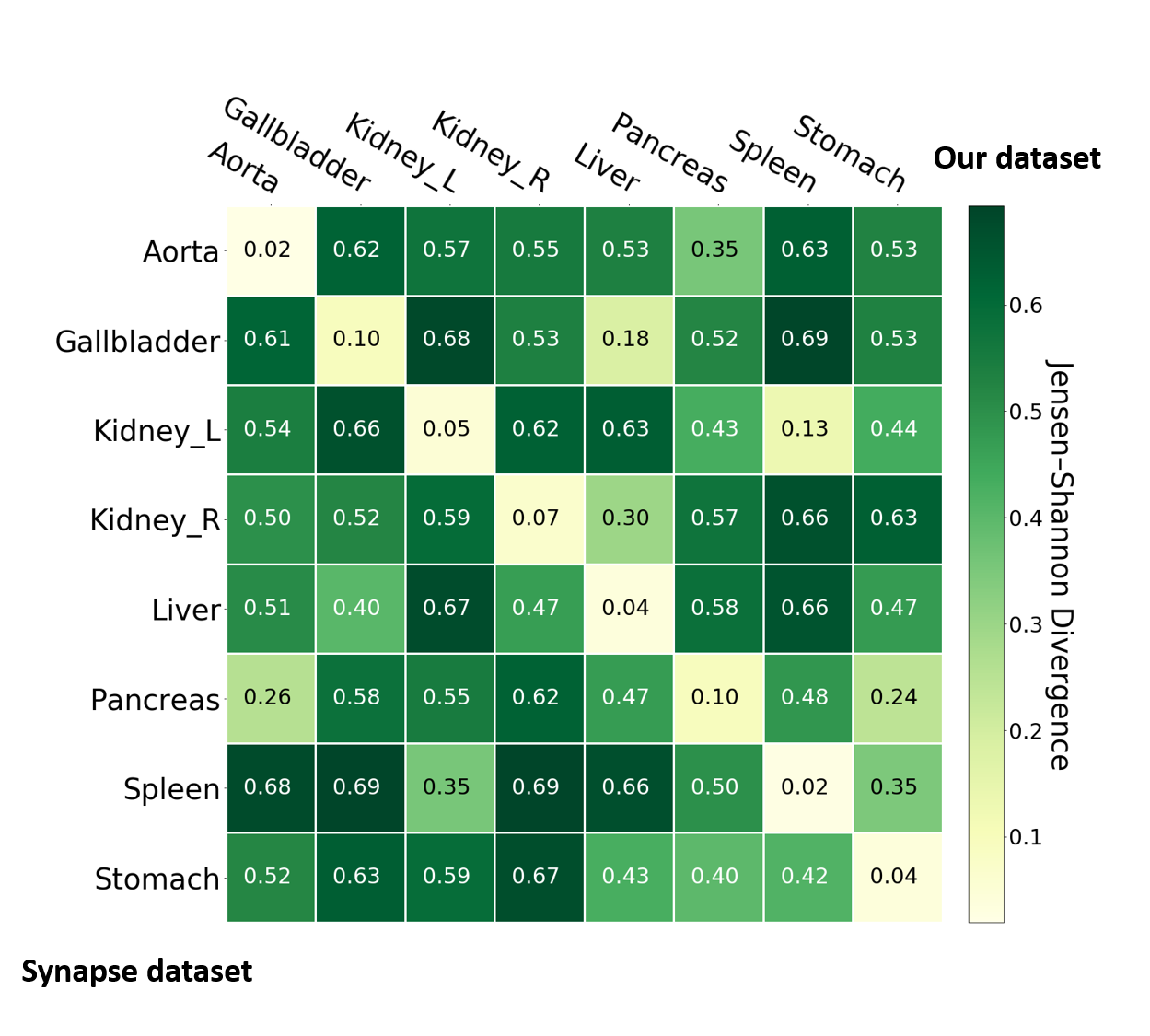}
\caption{We split each case into $ 3 \times 3 \times 3$ equally large patches and count the number of occurrences of an organ voxel appearing in each patch for Synapse and our dataset. We then calculate the Jensen–Shannon divergence of the two datasets. Smaller value means the row entry and the colume entry are closer.}
\label{fig:example}
\end{figure}

In this paper, we focus on unsupervised domain adaptation (UDA) for deviating acquisition scanners/protocols in 3D abdominal multi-organ segmentation on CT scans.
We propose a domain adaptive relational reasoning (DARR) by fully leveraging the organ location information. More concretely, the relative locations of organs remain stable in medical images \cite{Yao09statistical}. As an example shown in Fig.~\ref{fig:example}, we calculate the Jensen–Shannon divergence matrix of the location probability distribution of the 8 organs between Synapse dataset and our dataset. The co-occurrence of the same organ appearing in the same location is high. Such relational configuration is deemed as weak cues for segmentation task, which is easier to learn, and thus better in transfer \cite{Wei19iterative}. We aim at learning the spatial relationship of organs via recovering a CT scan from its shuffled patches, {\em{a.k.a}}, solving jigsaw puzzles. 
But, unlike previous methods which simply treated solving jigsaw puzzles as a regularizer in main tasks to mitigate the spatial correlation issue \cite{Carlucci19domain}, we also solve the jigsaw puzzle problem at test-time, based on one single test case presented. This can help us learn to adapt to a new target domain since the unlabeled test case provides us a hint about the distribution where it was drawn. It is worthwhile mentioning that this test-time relational reasoning process enables one model to adapt all.

To better learn the correlation of organs, we must guarantee that data from different domains have the same spatial resolution. Towards this end, we further propose a super-resolution network to jointly train with the segmentation network and the jigsaw puzzles, which can obtain high-resolution output from its low-resolution version. Since there exists a multiplicity of solutions for a given low-resolution voxel, we will show in the supplementary material that our super-resolution network has the capacity to learn better low-level features, {\em{i.e.}}, the deviation of voxels' Hounsfield Units within an organ is reduced, and that of inter-organ is enlarged.

Our proposed DARR performs test-time relative position training, which enjoys the following benefits: (1) establishing a naturally existed common constraint in medical images, so that it can easily adapt to unknown domains; (2) mapping data from different domain sites to the same spatial resolution and encouraging a more robust low-level feature for segmenting organs and learning organ relation; (3) free of re-training the network on source domain when adapting to new domains; and (4) outperforming baseline methods by a large margin, {\em{e.g.}}, with even over $29\%$ improvement in terms of mean DSC when adapting our model to multiple target datasets.

\section{Related Work}

(Unsupervised) Domain adaptation (UDA) has recently gained considerable interests in computer vision primarily for classification \cite{busto2018open}, detection \cite{zhu2019adapting}\cite{Ganin2015} and semantic segmentation \cite{zou2018unsupervised}\cite{bolte2019unsupervised}\cite{Sankaranarayanan_2018_CVPR}. A key principle of unsupervised domain adaptation is to learn domain invariant features by minimizing cross-domain differences either in feature-level or image-level \cite{sun2016deep}\cite{tzeng2014deep}. Inspired by the success of CycleGAN \cite{zhu2017unpaired} in unpaired image-to-image translation, many recent image adaptation methods are built upon modified CycleGAN frameworks to mitigate the impact of domain gap \cite{Murez_2018_CVPR}\cite{hoffman2017cycada}\cite{almahairi2018augmented}\cite{liu2017unsupervised}\cite{chang2018pairedcyclegan}. CyCADA \cite{hoffman2017cycada} poses unsupervised domain adaptation as style transfer with adversarial learning to close the gap in appearance between the source and target domains. Similar adversarial learning techniques are applied in cross-modality medical data \cite{chen2020unsupervised}\cite{dou2018unsupervised}\cite{Dou19pnp}\cite{kamnitsas2017unsupervised}\cite{joyce2018deep}. SIFA \cite{chen2020unsupervised} is among the latest GAN-based methods dedicated to adapt MR/CT cardiac and multi-organ segmentation networks, which conducts both image-level and feature-level adaptations with a shared encoder structure. 

More recently, there have been multiple self-training/pseudo-label based methods for unsupervised domain adaptation \cite{zou2018unsupervised}\cite{Zou_2019_ICCV}\cite{busto2018open}\cite{inoue2018cross}\cite{zhou2019semi}. \cite{zhou2019semi} proposes a semi-supervised 3d abdominal multi-organ segmentation by first training a teacher model in source dataset in a fully-supervised manner and compute the pseudo-labels on the target dataset. Then a student model is trained on the union of both datasets. However, domain shift is not delicately addressed in this method, thus it hampers its usage on domain adaptation tasks. Another important class for unsupervised domain adaptation is based on self-supervised learning \cite{chen2019self}\cite{Carlucci19domain}. The key challenge for self-supervised learning is identifying a suitable self supervision task. Patch relative positions \cite{doersch2015unsupervised}, local context \cite{pathak2016context}, color \cite{zhang2017split}, jigsaw puzzles \cite{Wei19iterative} and even recognizing scans of the same patient \cite{jamaludin2017self} have been used in self-supervised learning. In this paper, we aim at learning the spatial relationship of organs via recovering a CT scan from its shuffled patches.

\begin{figure*}[t]
\centering
\includegraphics[width=1\linewidth]{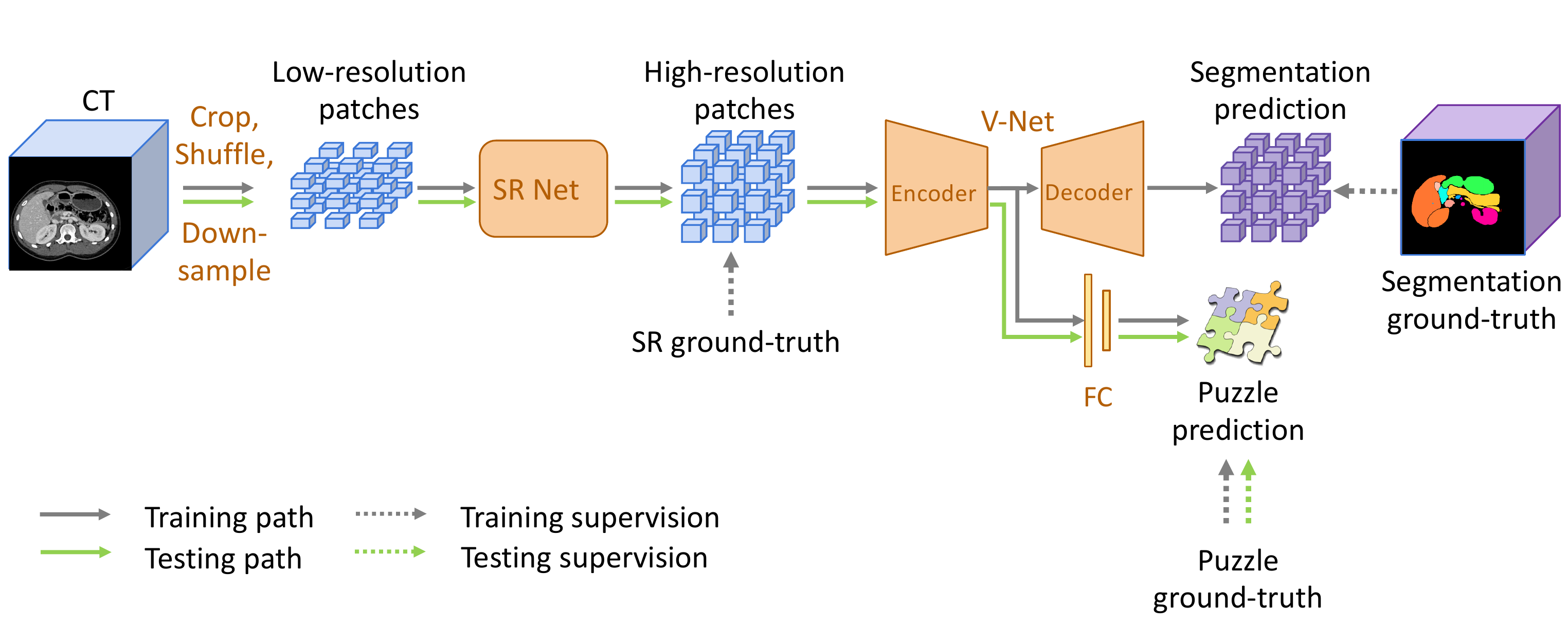}
\caption{An overview of our model. Our framework consists of three components, a super-resolution network that upsamples low-resolution images to high resolution, a standard V-Net that performs the segmentation task, and a puzzle module to learn the spatial relations among patches.}
\label{fig:overview}
\end{figure*}

\section{Method}

\subsubsection{Problem definition.}
Our goal is to develop a framework that enables machine learning trained on one source domain to adapt to multiple target domains during testing. An overview of our architecture is shown in Fig. \ref{fig:overview}. Our framework consists of three components, a super-resolution network that upsamples low-resolution images to high resolution, a standard V-Net \cite{vnet} that performs the segmentation task and a puzzle module to learn the spatial relations among patches. We adopt the generator network from \cite{sr} with the subpixel upsampling method as our super-resolution module and we will show the details of the puzzle module in the following section.

We first define some notations in the paper. We parametrize the super-resolution network as $\theta_{sr}$, the encoder part of V-Net as $\theta_{en}$ - which is shared by the puzzle module, the decoder part of V-Net as $\theta_{de}$ and the puzzle module as $\theta_{p}$. 
Suppose we are partitioning an image $\textbf{I}$ into $W \times H \times L$ patches where each patch can be denoted as $\textbf{i}_{xyz}$. The index $x \in \{1,...,W\}$, $y \in \{1,...,H\}$ and $z \in \{1,...,L\}$ indicate the original relative location from which the patches are cropped. Then each patch $\textbf{i}_{xyz}$ can be associated with a unique label $l_{xyz}$ following the row-major policy $l_{xyz} = x + Wy + WHz$ that serves as the ground truth in the jigsaw puzzle task. 
We use $\{\textbf{i}_{a}\}$ to indicate a random permutation of the patch set $\{\textbf{i}_{xyz}\}$, and the label $l_{xyz}$ for each patch $\textbf{i}_{xyz}$ is also permutated the same way, denoted as $l_a$, where $a \in \{1, ..., WHL\}$. 
\subsubsection{Training stage.}
Our network can be trained end-to-end, with one loss from each module. To train the super-resolution network, we squeeze the image patch $\textbf{i}_a$ to a smaller size $\textbf{i}_a^\prime$ and minimize a mean square loss $L_{sr}(\textbf{i}_a^\prime, \textbf{i}_a; \theta_{sr})$ which makes the output patch as close as possible to the original patch. 
The segmentation network produces a cross-entropy loss $L_{seg}(\textbf{i}_a, \textbf{g}_a; \theta_{sr}, \theta_{en}, \theta_{de})$, where $\textbf{g}_a$ is the ground truth segmentation mask. The third loss, $L_p(\textbf{i}_a, l_a; \theta_{sr}, \theta_{en}, \theta_p)$, is given by the puzzle task that classifies the correct location of the patches. 
Note that the former two losses $L_{seg}$ and $L_{sr}$ are only trained on the training dataset, while the puzzle loss $L_p$ can be utilized on both training and testing set because it does not require any manually labeled data. 

Overall, we can obtain the optimal model through
\begin{equation}
    \argmin{\theta_{sr}\ \theta_{en}\ \theta_{de}\ \theta_{p}} L_{seg}(\textbf{i}_a, \textbf{g}_a; \theta_{sr}, \theta_{en}, \theta_{de}) + \lambda_{sr}L_{sr}(\textbf{i}_a^\prime, \textbf{i}_a; \theta_{sr}) + \lambda_p L_p(\textbf{i}_a, l_a; \theta_{sr}, \theta_{en}, \theta_p),
\end{equation}
where $\lambda_{sr}$ and $\lambda_p$ are loss weights.
\subsubsection{Adaptive testing.}
During testing, our goal is to adapt our feature extractor to the target domain, or a target image, through optimizing the self-supervised learning task. By minimizing the puzzle loss $L_p$ on the testing data for a few iterations, the feature extractor is able to reason about the spatial relations among organs, and thus improving the performance on the unseen target domain. 
\subsubsection{Jigsaw puzzle solver.}
Medical images share a strong spatial relationship that organs are organized in specific locations inside the body with similar relative scales. With this prior knowledge, it is natural to investigate a self-supervised learning task that solves for the relative locations given arbitrarily cropped 3D patches. We select a Jigsaw Puzzle Solver in our case, as it has been proven to be helpful in initializing 3D segmentation models \cite{Wei19iterative}. During training, the permuted set of patches $\{\textbf{i}_a^\prime\}$ are passed through the super-resolution network and the shared feature extractor (the encoder part) of V-Net to generate corresponding features, denoted as $\{\textbf{f}_a\}$. Following the previous work \cite{Wei19iterative}, all features are then flattened into 1D vectors and concatenated together according to the permuted order, forming a long vector. After two fully-connected layers, the puzzle module outputs a vector in size $(WHL)^2$, which can be reshaped into a matrix of size $WHL \times WHL$. We apply a softmax function on each row so that each row $a$ of the matrix indicates the probability of patch $\{\textbf{i}_a^\prime\}$ belonging to the $WHL$ locations. We use negative log-likelihood loss as the puzzle loss in our model.

\section{Experiments}
\subsubsection{Datasets.}
We train the proposed DARR model on our high-resolution multi-organ dataset with 90 cases and adapt it to five different public medical datasets, including 1) multi-organ dataset: Synapse dataset\footnote{\url{https://www.synapse.org/\#!Synapse:syn3193805/wiki/217789}} (30 cases); and 2) 4 single-organ datasets \cite{Simpson2019}: Spleen (41 cases), Liver (131 cases), Pancreas (282 cases) and NIH Pancreas dataset\footnote{\url{https://wiki.cancerimagingarchive.net/display/Public/Pancreas-CT}} (82 cases). 
\textbf{For the Synapse dataset}, we evaluate on 8 abdominal organs, including Aorta, Gallbladder, Kidney (L), Kidney (R), Liver, Pancreas, Spleen and Stomach, which are also annotated in our multi-organ dataset. \textbf{For all other datasets}, we directly evaluate on the target organ.
Each dataset is randomly split into 80\% of training data and 20\% of testing data. Note that unlike other domain generalization methods which use data from the target domain for training, DARR only sees target domain data during testing. We use Dice-S{\o}rensen coefficient (DSC) as the evaluation metric. For each target dataset, we report the average DSC of all cases.

\subsubsection{Implementation details.}
We set puzzle-related hyperparameters $W = H = L = 3$ in all experiments, which leads to a puzzle composed of 27 patches. 
The loss-related weights are set as $\lambda_{sr} = 30$ and $\lambda_p = 0.1$, which are consistent across all experiments. 

We use 3D V-Net as our backbone architecture, which is initialized with a standard V-Net pre-trained on our in-house dataset. The puzzle module shares the same encoder with the segmentation branch with an additional classification head. We use two fully-connected layers to generate puzzle prediction. 
The whole network is then finetuned with Adam solver for another 40000 iterations with the batch size of 1 and a learning rate of 0.0003. Each patch has size $64 \times 64 \times 64$, and is squeezed into size $64 \times 64 \times 16$ before feeding into DARR.  

For each target dataset, we further train a supervised V-Net model with their ground-truth labels and test directly on the same target dataset. These results serve as our upper bound performance and can be used to calculate the performance degradation for source-to-target adaptation. 

During testing, the DARR is first finetuned with a puzzle module only from each target image for 30 iterations with a learning rate of 1e-5 and SGD solver. Then we fix the network parameters and output the segmentation results via a forward pass through the segmentation branch. After predicting one target image, the model is rolled back to the original model, and the above test-time jigsaw puzzle training is repeated for the next target image. No further post-processing strategies are applied.

\begin{table}[t!]
\centering
\begin{adjustbox}{width=1\textwidth}
\begin{tabular}{@{}l@{\hspace{6mm}}c@{\hspace{4mm}}c@{\hspace{4mm}}c@{\hspace{4mm}}c@{\hspace{4mm}}c@{\hspace{4mm}}c@{}}
\toprule
           & Synapse & MSD Liver & MSD Pancreas & MSD Spleen & NIH Pancreas & Average \\ 
\midrule
Lower Bound  & 52.42 & 88.10   & 24.42     & 43.31   & 10.55   & 43.76  \\
SIFA\cite{chen2020unsupervised}  & 62.33   & 91.58   & 24.68   &   89.37   & 18.76   &  57.35     \\
VNET-Puzzle    & 60.99 & 89.28 & 36.22 & 60.25 & 31.32 & 55.61 \\
VNET-SR & 61.27 & 88.83 & 47.09 & 80.82 & 35.71 & 62.74 \\
DARR         & 69.77 & 92.33   & 56.12      & 88.61    & 59.96   & 73.36  \\
\midrule
Upper Bound & 68.81 & 91.55  & 72.56     & 86.43    & 86.37   & 81.14  \\
\bottomrule
\end{tabular}
\end{adjustbox}
\caption{Domain generalization results (DSC $\%$) on all five target datasets. 
}
\label{table:main_result}
\end{table}

\subsubsection{Results and discussions.}
We compare our DARR with state-of-the-art methods, \emph{i.e.}, GAN-based methods~\cite{chen2020unsupervised}, self-learning-based methods~\cite{zhou2019semi}, and meta-learning-based methods~\cite{dou2019domain}. 
The performance comparison on different datasets is shown in Table~\ref{table:main_result}. 
To measure the performance gain after adaptation, we also provide results trained on our in-house dataset and tested directly on the target datasets without DARR (denoted as ``Lower Bound'' in Table~\ref{table:main_result}).
We observe that our method improves Lower Bound results by 29.60\% on average and outperforms all other methods by a large margin. It is worth noting that our method even outperforms Upper Bound results on Synapse dataset, MSD Liver dataset, and MSD Spleen dataset, without using any target domain data in training. 
This result indicates that our method, which captures spatial relations among organs, is able to bridge the domain gap between multi-site data. 
\paragraph{Comparison with self-learning.} Following~\cite{zhou2019semi}, we first train a teacher model on our multi-organ dataset in a fully-supervised manner and compute the pseudo-labels on the Synapse dataset. Then a student model is trained on the union of both datasets.
By evaluating the Synapse dataset, we find that the student model yields a lower segmentation performance than that of the teacher model. This indicates that simply using self-learning may not effectively distill information from data of a different source site.
\paragraph{Comparison with Meta-learning model-agnostic learning methods.} The MASF~\cite{dou2019domain} splits the source domain into multiple non-overlapping training and testing sets and trains a meta-learning model-agnostic model viewing the smaller set as different tasks. It also utilizes delicately designed losses to align intra-class features and separate inter-class features. Nevertheless, MASF does not transfer well from the source domain to the target domains.
It is only able to transfer large organs like the liver and stomach while performs poorly in detecting the other small organs. This further confirms that the domain gaps among datasets are substantial, especially in multi-organ segmentation and cannot be easily solved by Meta-learning methods.

\begin{figure}[!t]
\centering
\includegraphics[width=0.85\linewidth]{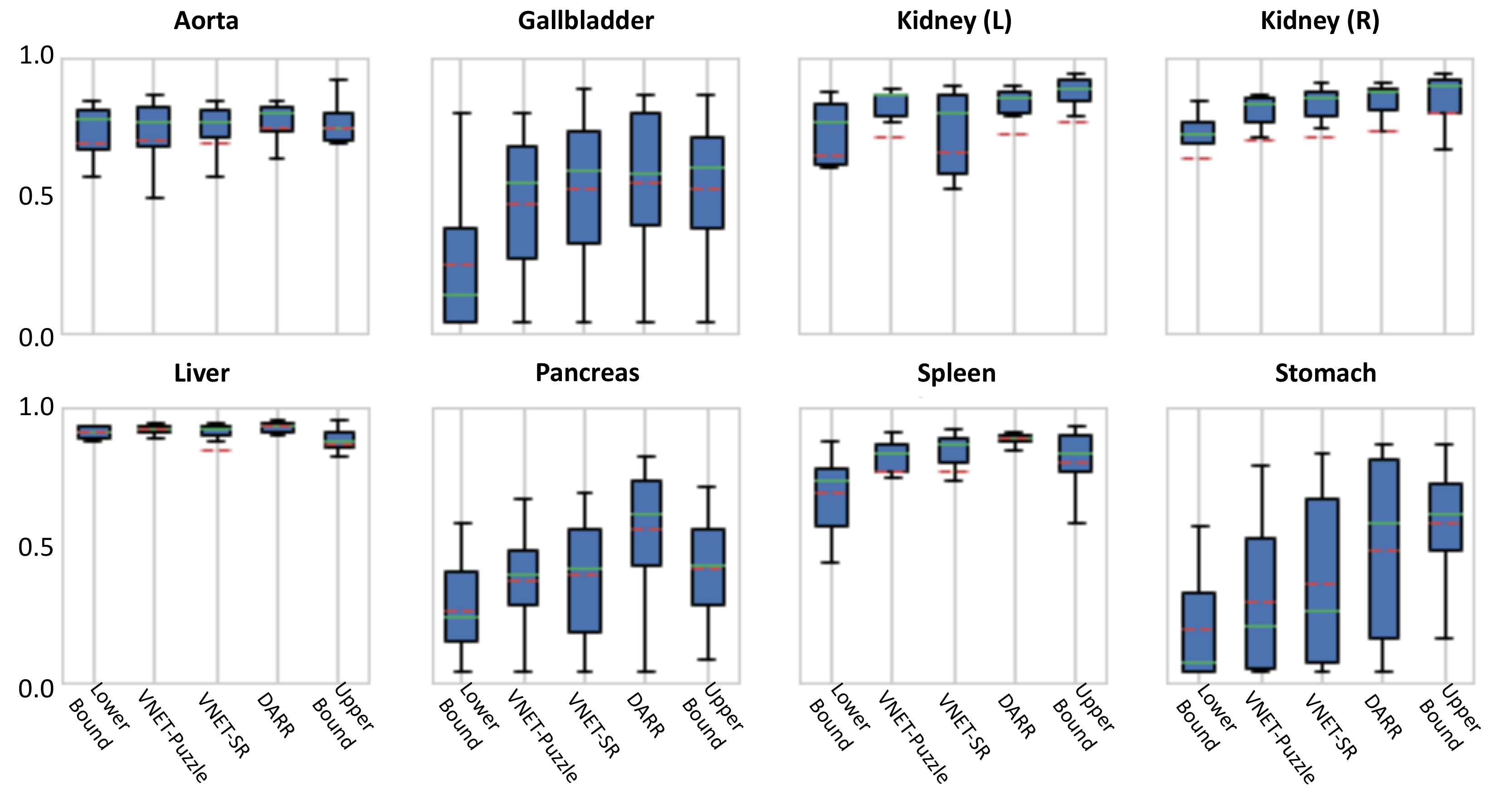}
\caption{Ablation study on key components of DARR with Synapse dataset.}
\label{Fig:boxplot}
\end{figure}

\paragraph{Comparison with GAN-based methods.} The SIFA \cite{chen2020unsupervised} is dedicated to adapt MR/CT cardiac and multi-organ segmentation networks. It conducts both image-level and feature-level adaptations based on a modifed CycleGAN \cite{zhu2017unpaired}. 
We use the generated target domain images and their corresponding ground truth in the source domain to train a target segmentation network. Here we apply DeepLab-v2 \cite{chen2017deeplab} for training the segmentation network after image adaptation of SIFA. From Table \ref{table:main_result} we can see that our VNET-SR already outperforms SIFA and achieves inspiring results with average Dice increased to 62.74\%. Our full DARR recovers the performance degradation further by an average of $29.6\%$ compared with lower bound, and outperforms SIFA by a significant margin (only $13.59\%$ by SIFA), which shows the superior performance of DARR.

\subsubsection{Ablation Study.}
\begin{figure}[t]
\begin{adjustbox}{max width=\textwidth}{
	\begin{minipage}[c][1\width]{
	   0.2\textwidth}
	   \centering
	   \includegraphics[width=1.0\textwidth]{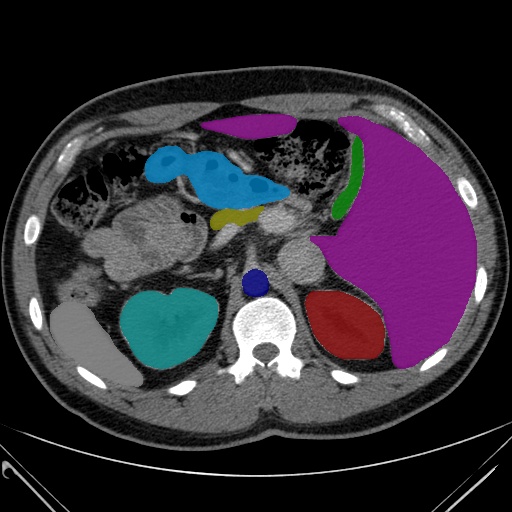}
	\end{minipage}
 \hfill 	
	\begin{minipage}[c][1\width]{
	   0.2\textwidth}
	   \centering
	   \includegraphics[width=1.0\textwidth]{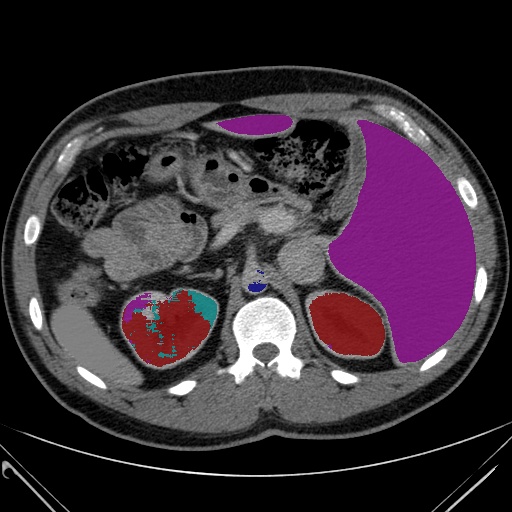}
	\end{minipage}
 \hfill	
	\begin{minipage}[c][1\width]{
	   0.2\textwidth}
	   \centering
	   \includegraphics[width=1.0\textwidth]{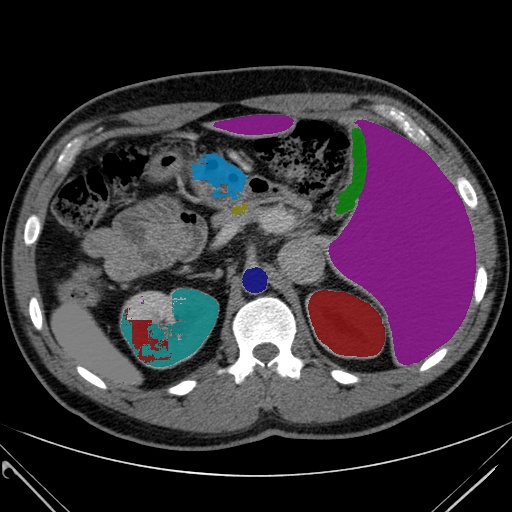}
	\end{minipage}
 \hfill	
	\begin{minipage}[c][1\width]{
	   0.2\textwidth}
	   \centering
	   \includegraphics[width=1.0\textwidth]{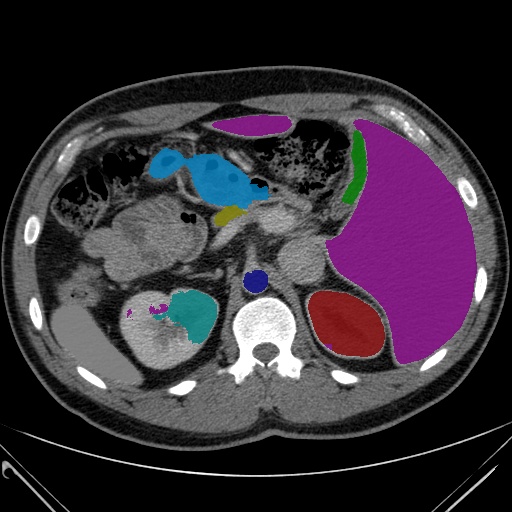}
	\end{minipage}
 \hfill	
	\begin{minipage}[c][1\width]{
	   0.2\textwidth}
	   \centering
	   \includegraphics[width=1.0\textwidth]{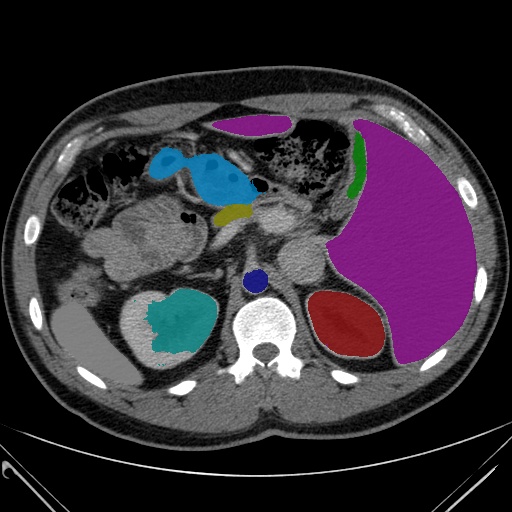}
	\end{minipage}
}\end{adjustbox}
\caption{Qualitative comparison of different approaches. From left to right: a) Ground Truth, b) Lower Bound, c) VNET-Puzzle, d) VNET-SR, e) DARR. Our method can successfully distinguish between left kidney and right kidney after learning their spatial relations.}
\label{fig:ablation}
\end{figure}

In this section, we evaluate how each component contributes to our model. 
We compare different variants of our method (using V-Net as the backbone model): 1) \textbf{VNET-Puzzle}, which integrates an additional puzzle module to adaptively learn the spatial relations among image patches; 2) \textbf{VNET-SR}, which employs a super-resolution module before the segmentation network; and 3) our proposed \textbf{DARR} with both the puzzle module and the super-resolution module applied. As can be seen from Table~\ref{table:main_result}, compared with Lower Bound (which simply uses bilinear upsampling strategies to overcome the resolution divergence among datasets), VNET-SR consistently achieves performance gains on all 5 different target datasets.
Especially, for more challenging datasets, the performance improvement can be significant and substantial, \emph{e.g.}, $8.86\%$ on the Synapse dataset, $22.68\%$ on the MSD Pancreas dataset and $25.16\%$ on the NIH Pancreas dataset.
This finding indicates the efficacy of our super-resolution module for handling the resolution differences among multi-site data. 
In addition, VNET-Puzzle also consistently outperforms Lower Bound by a large margin, \emph{e.g.}, $60.99\%$ vs. $52.42\%$ for the Synapse dataset, $36.22\%$ vs. $24.42\%$ for the MSD Pancreas dataset, and $10.55\%$ vs. $31.32\%$ for the NIH Pancreas dataset.
Equipped with both the puzzle module and the super-resolution module, our DARR can even lead to additional performance gains compared with VNET-Puzzle and VNET-SR. For instance, we observe an improvement of $17.36\%$ on the Synapse dataset, $31.70\%$ on thee MSD Pancreas dataset, and over $40\%$ on both the MSD Spleen dataset and the NIH Pancreas dataset.
We also provide component comparison results in box plots (see Fig.~\ref{Fig:boxplot}) for the Synapse dataset, which suggests a general statistical improvement among all tested organs. To further demonstrate the efficacy of the proposed DARR, a qualitative comparison is illustrated in Fig.~\ref{fig:ablation}, where the spatial location of both kidneys is successfully identified by DARR.

\section{Conclusions}
We proposed an unsupervised domain adaptation method to generalize 3D multi-organ segmentation models to medical images collected from different scanners and/or protocols (domains). This method, named Domain Adaptive  Relational Reasoning, is inspired by the the fact that the spatial relationship between internal structures in medical images are relatively fixed. We formulated the spatial relationship by solving jigsaw puzzles and utilized two schemes, \emph{i.e.}, spatial resolution standardisation and test-time jigsaw puzzle training, to guarantee its transferability to multiple domains. Experimental results on five public datasets demonstrate the superiority of our method.

\section*{Acknowledgement}
We especially Chen Wei for her valuable discussions and ideas. This work was supported by the Lustgarten Foundation for Pancreatic Cancer Research.


%
%

\bibliographystyle{splncs04}
\bibliography{paper386}
\end{document}